\documentclass[runningheads]{llncs}
\usepackage[T1]{fontenc}
\usepackage{graphicx}
\usepackage{cite}
\usepackage{amsmath,amssymb,amsfonts}
\usepackage{algorithmic}
\usepackage{algorithm}
\usepackage{graphicx}
\usepackage{textcomp}
\usepackage{xcolor}
\usepackage{float}
\restylefloat{table}
\usepackage{lscape}
 \usepackage[numbers]{natbib}
\usepackage{graphicx}
\usepackage{caption}
\usepackage{subcaption}
\usepackage{booktabs,makecell}
\usepackage{multirow}
\usepackage{pdflscape}
\usepackage{caption}
\usepackage{hyperref}
\usepackage{rotating}
\usepackage[export]{adjustbox}
\usepackage{comment}
\usepackage{amsmath}
\usepackage{xcolor}
\usepackage{tikz}
\usetikzlibrary{shapes,arrows}
\usepackage{placeins}
\usepackage{graphics}
\usepackage{xcolor}
\usepackage{mathrsfs}

\def\BibTeX{{\rm B\kern-.05em{\sc i\kern-.025em b}\kern-.08em
    T\kern-.1667em\lower.7ex\hbox{E}\kern-.125emX}}
\begin{document}
\title{Advancing RVFL networks: Robust classification with the HawkEye loss function}

\author{Mushir Akhtar\inst{1} \and Ritik Mishra\inst{1}  \and M. Tanveer\inst{1}\thanks{\noindent Corresponding Author} \and Mohd. Arshad\inst{1}
}
\institute{Indian Institute of Technology Indore, Simrol, Indore, India 
\email{\{phd2101241004,phd2301241003,mtanveer,arshad\}@iiti.ac.in}}
\authorrunning{Akhtar et al.}
\maketitle       

\begin{abstract}
Random vector functional link (RVFL), a variant of single-layer feedforward neural network (SLFN), has garnered significant attention due to its lower computational cost and robustness to overfitting. Despite its advantages, the RVFL network's reliance on the square error loss function makes it highly sensitive to outliers and noise, leading to degraded model performance in real-world applications. To remedy it, we propose the incorporation of the HawkEye loss (H-loss) function into the RVFL framework. The H-loss function features nice mathematical properties, including smoothness and boundedness, while simultaneously incorporating an insensitive zone. Each characteristic brings its own advantages: 1) Boundedness limits the impact of extreme errors, enhancing robustness against outliers; 2) Smoothness facilitates the use of gradient-based optimization algorithms, ensuring stable and efficient convergence; and 3) The insensitive zone mitigates the effect of minor discrepancies and noise. Leveraging the H-loss function, we embed it into the RVFL framework and develop a novel robust RVFL model termed H-RVFL. Notably, this work addresses a significant gap, as no bounded loss function has been incorporated into RVFL to date. The non-convex optimization of the proposed H-RVFL is effectively addressed by the Nesterov accelerated gradient (NAG) algorithm, whose computational complexity is also discussed. 
The proposed H-RVFL model's effectiveness is validated through extensive experiments on $40$ benchmark datasets from UCI and KEEL repositories, with and without label noise. The results highlight significant improvements in robustness and efficiency, establishing the H-RVFL model as a powerful tool for applications in noisy and outlier-prone environments. The supplement material and the source code for the proposed H-RVFL are publicly available at \url{https://github.com/mtanveer1/H-RVFL}.

\end{abstract}

\section{Introduction}
In real-world scenarios, the pervasive impact of outliers and noise on machine learning techniques poses significant challenges to model robustness and accuracy \cite{blazquez2021review, smiti2020critical}. These anomalies disrupt the learning process by introducing irregularities that obscure the true data patterns, leading to the development of erroneous models \cite{lin2002fuzzy}. Traditional models often employ unbounded loss functions, like squared error loss or hinge loss, which are particularly susceptible to the influence of outliers and noise. These loss functions amplify the effects of aberrant data points, skewing the model’s performance and resulting in suboptimal outcomes.
\par
The random vector functional link (RVFL) network \cite{pao1994learning} is an efficient and versatile neural network model successfully applied across various domains, including time-series prediction \cite{gao2021walk}, Alzheimer's diagnosis \cite{malik2022graph, sharma2022conv}, Schizophrenia diagnosis \cite{varaprasad2024effective}, and so forth.
Its ability to handle high-dimensional data and robustness to overfitting make it a popular choice for many real-world applications. RVFL is a single-layer feedforward neural network (SLFN) that distinguishes itself through its use of randomized weights and biases for the input-to-hidden layer connections, while the output layer weights are determined through a closed-form solution. Due to the fixed random weights and biases in the hidden layer, the RVFL network significantly reduces the computational cost associated with training compared to traditional neural networks. The combination of randomization and direct links helps improve generalization performance by capturing both linear and non-linear relationships in the data \cite{zhang2016comprehensive, malik2023random}.
\par
Despite the numerous advantages of RVFL, ongoing research is continually improving its efficiency and robustness. \citet{xu2017kernel} proposed a kernel-based RVFL model for effective spatiotemporal modeling. \citet{zhao2021consistency} developed a multi-view multi-label learning framework based on RVFL to address consistency and diversity issues.
\citet{sajid2024neuro} combined the neuro-fuzzy system \cite{de2020fuzzy} with RVFL, creating a neuro-fuzzy RVFL that utilizes the IF-THEN approach to make human-like decisions and enhance RVFL robustness. \citet{ganaie2024graph} proposed the graph-embedded intuitionistic fuzzy RVFL to address class imbalance problems. Graph embedding is utilized to maintain the inherent topological structure of datasets, while intuitionistic fuzzy theory is used to manage uncertainty and imprecision in the data.
\par
Alternatively, while there have been several advancements in RVFL, there remains room for improvement. One key area is to focus on its loss function. The majority of existing works on RVFL have employed the square error loss function, which is highly sensitive to outliers. In the presence of outliers, the square error loss function can lead to extreme sensitivity, as the quadratic penalty applied to large errors disproportionately affects the loss function, resulting in an overemphasis on minimizing these outliers \cite{fu2023robust}. As a result, the model’s performance can degrade, leading to suboptimal results, especially in noisy scenarios \cite{akhtar2024advancing}. To address this issue to some extent, authors in \cite{hazarika2022random} and \cite{xie2023huber} incorporated the Huber loss function into RVFL. Huber loss exhibits a quadratic behavior for minor deviations and transitions to a linear regime for larger discrepancies, thereby attenuating the impact of outliers compared to the squared error loss function. However, it still increases unboundedly, allowing extremely large errors to exert a disproportionate influence, potentially undermining the model’s robustness and generalization capabilities. To improve the robustness of machine learning algorithms, recently various bounded loss functions have been developed, including RoBoSS loss \cite{10685140}, wave loss \cite{akhtar2024advancing}, HawkEye loss \cite{akhtar2024hawkeye}, and so forth. These functions implement a maximum threshold and halt the escalation of losses beyond a specific point, thereby restricting the influence of outliers more effectively. Despite these advancements, no bounded loss function has been integrated into the RVFL framework to date, presenting a significant opportunity for research advancement.
\par
To address this gap, this paper aims to develop a robust RVFL model by incorporating a bounded loss function. Among the various bounded loss functions, we select the HawkEye loss (H-loss) \cite{akhtar2024hawkeye} for its dual advantages of an insensitive zone and inherent smoothness properties. The insensitive zone allows the H-loss to tolerate small errors, which is particularly beneficial in mitigating the impact of noise and minor discrepancies within the data. This tolerance ensures that the model does not overreact to insignificant deviations, thereby enhancing its robustness. Additionally, the smoothness characteristic of the H-loss function enables the utilization of gradient-based optimization techniques. The smoothness guarantees the existence of well-defined gradients, facilitating the application of efficient and reliable optimization methods. By integrating the H-loss function into the RVFL framework, we propose a novel robust RVFL model named H-RVFL. The non-convexity issue associated with the H-RVFL is effectively resolved by leveraging the Nestrov accelerated gradient (NAG) algorithm. The key highlights of this paper can be encapsulated as follows:
\begin{enumerate}
    \item We address a significant gap in RVFL research by incorporating the HawkEye loss (H-loss) function, the first bounded loss function to be embedded into the RVFL framework. This integration aims to improve the robustness of RVFL against outliers and noise, which are prevalent in real-world datasets.
    \item We construct a novel robust RVFL model for binary classification, termed H-RVFL, leveraging the unique properties of the H-loss function, including boundedness, smoothness, and an insensitive zone. These properties collectively improve the model's resilience to extreme errors and reduce the impact of minor discrepancies and noise.
    \item We use the NAG algorithm to effectively address the non-convex optimization problem of the proposed H-RVFL model. It provides faster convergence rates, efficient memory utilization, and reduced oscillations during training. The computational complexity of the NAG algorithm is also discussed.
    \item We validate the proposed H-RVFL model through extensive experiments on $40$ benchmark UCI and KEEL datasets of various domains, both with and without label noise. The results undeniably showcase the superiority of the proposed H-RVFL model in comparison to the evaluated models.
\end{enumerate}
The rest of the paper is structured as follows: Section 2 provides a review of the RVFL framework and the notations used throughout the paper. Section 3 discusses the H-loss function and its characteristics, and presents the formulation of the proposed H-RVFL model. Section 4 presents the experimental results. Finally, the concluding remarks are discussed in Section 5.

\section{Related Work}
In this section, we discuss the square error loss function and its drawbacks. Further, we provide the structure and mathematical formulation of the standard RVFL network in Section S.I of the supplement file.
\subsection{Squared error loss function}
The squared error loss function, also known as the mean squared error (MSE) when averaged over all data points, is one of the most commonly used loss functions in machine learning \cite{wang2020comprehensive}. It measures the square of the difference between the observed and predicted values. Mathematically, for a set of predictions \( \hat{y}_i \) and corresponding true values \( y_i \), the squared error loss is defined as:
\begin{align}
\mathcal{L}_{SE} = \frac{1}{n} \sum_{i=1}^{n} (y_i -\hat{y}_i)^2,
\end{align}
where \( n \) is the number of observations. Despite its widespread use, the squared error loss function has several notable drawbacks, particularly when applied to real-world data:
\begin{enumerate}
    \item It is sensitive to outliers. Since the loss function squares the error, any large deviations between the predicted and actual values are disproportionately penalized. This means that a few extreme outliers can heavily influence the model, leading to suboptimal parameter estimates and degraded performance.
    \item Due to its high sensitivity to outliers, the squared error loss is not robust. In practical scenarios, datasets often contain noise and outliers, which can severely impact the performance of models trained with MSE.
    \item It penalizes all deviations from the true values, which may lead to overfitting in noisy environments.
\end{enumerate}
Addressing these drawbacks often involves alternative loss functions, such as Huber loss \cite{wang2020comprehensive}, which balances the robustness to outliers with the advantages of squared error loss, or bounded loss functions \cite{akhtar2024hawkeye, akhtar2024advancing} that limit the impact of extreme errors.
\section{Proposed work}
In this section, we begin by discussing the HawkEye loss function and its distinguishing characteristics. Following this, we embed the HawkEye loss function into the RVFL and construct a novel robust RVFL model for binary classification named H-RVFL. Finally, we employ the NAG algorithm to optimize the H-RVFL model.
\subsection{HawkEye loss (H-loss) function} 
\begin{figure*}
\centering
    \subcaptionbox{     \label{fig:first}} { %
      \includegraphics[width=0.48\textwidth,keepaspectratio]{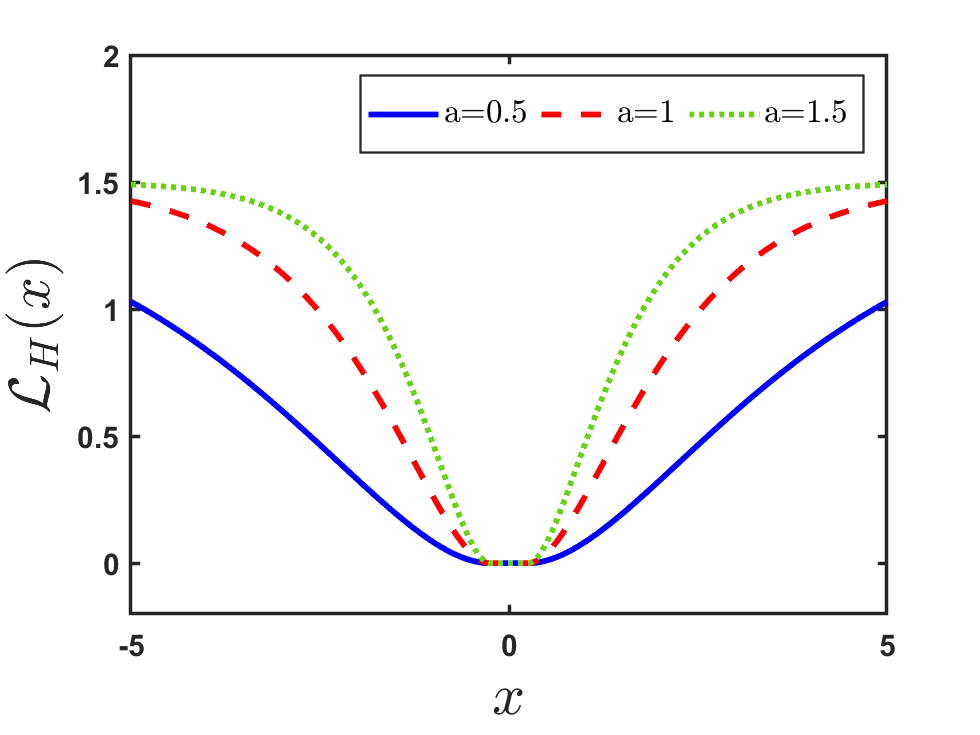}}
      \hfill
      \subcaptionbox{   \label{fig:second}} { %
      \includegraphics[width=0.48\textwidth,keepaspectratio]{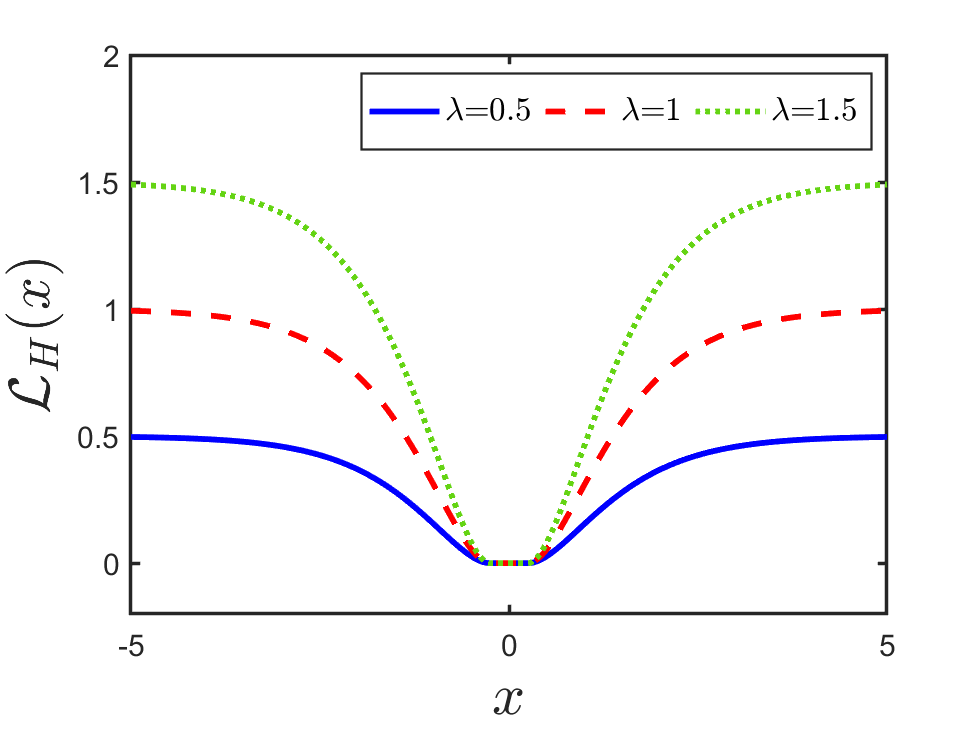}}
\\
      \subcaptionbox{  \label{fig:third}} { %
      \includegraphics[width=0.48\textwidth,keepaspectratio]{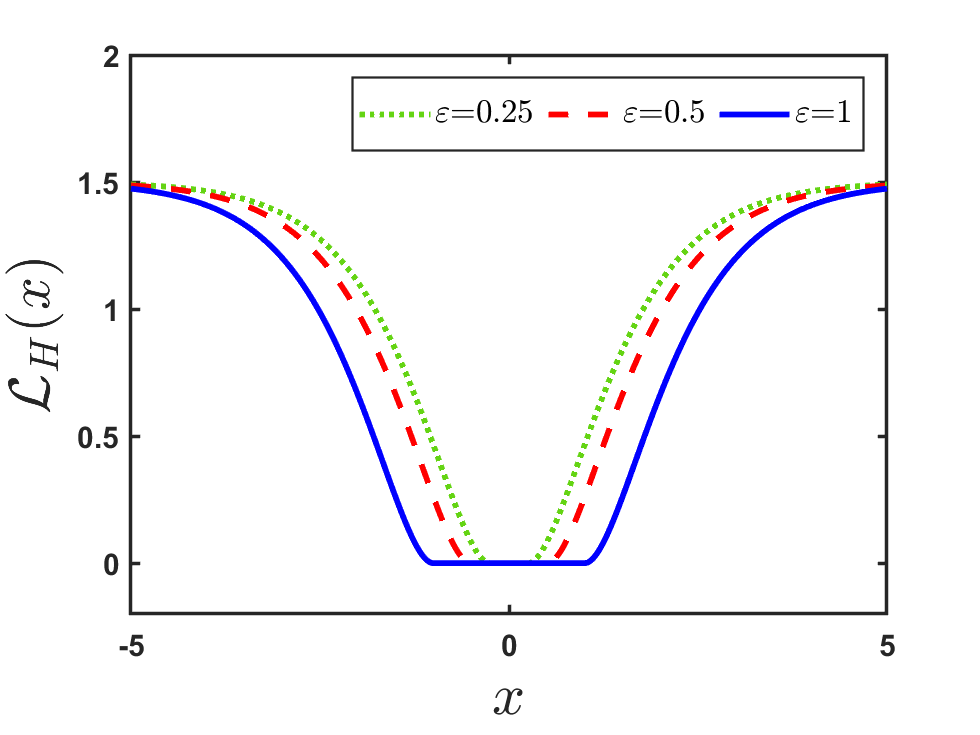}}
      \hfill     
      \subcaptionbox{  \label{fig:four}} { %
      \includegraphics[width=0.48\textwidth,keepaspectratio]{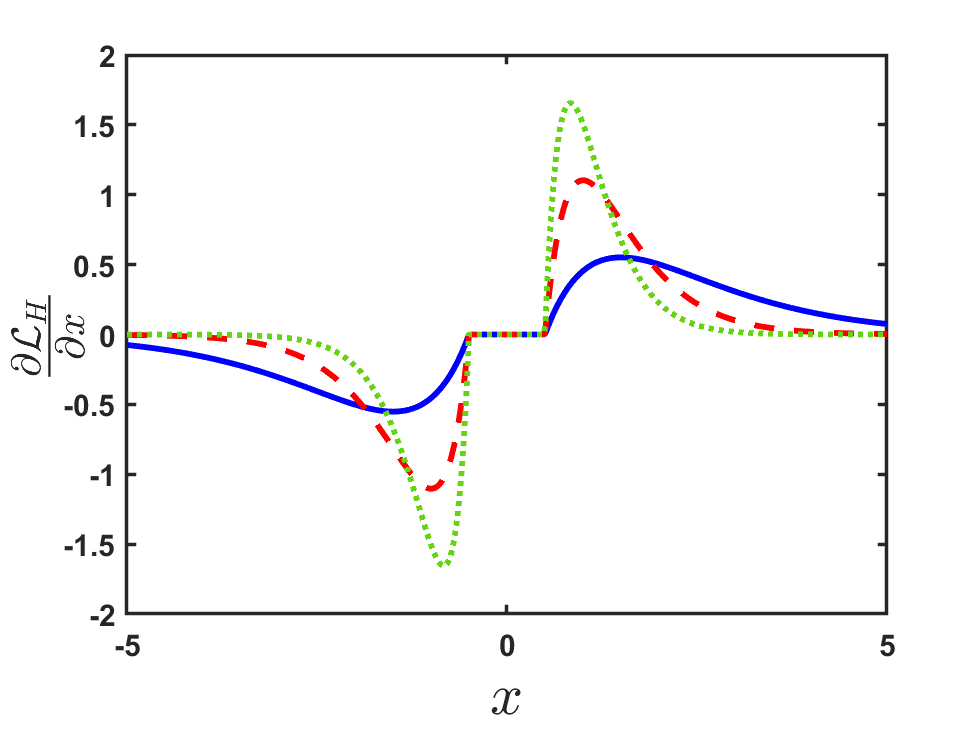}}
\\
      \subcaptionbox{   \label{fig:fifth}} { %
      \includegraphics[width=0.48\textwidth,keepaspectratio]{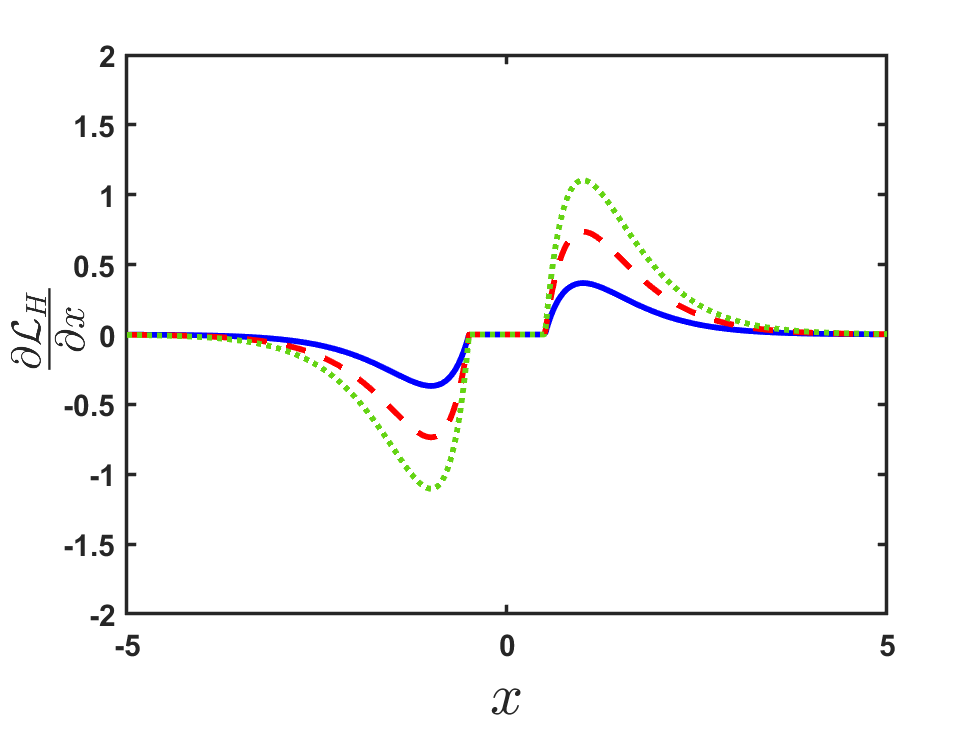}}
      \caption{Visualization of the H-loss function and its gradient. Subfigures (a), (b), and (c) demonstrate the H-loss behavior for varying values of \( a \) (with \( \lambda \) and \( \varepsilon \) fixed), \( \lambda \) (with \( a \) and \( \varepsilon \) fixed), and \( \varepsilon \) (with \( \lambda \) and \( a \) fixed), respectively. Subfigures (d) and (e) illustrate the gradient of the H-loss for different values of \( a \) (with \( \lambda \) and \( \varepsilon \) fixed) and \( \lambda \) (with \( a \) and \( \varepsilon \) fixed), respectively, emphasizing the function’s sensitivity to parameter variations and its behavior during optimization.} 
    \label{fig:Loss-Figues}
 \end{figure*}

Recently, \citet{akhtar2024hawkeye} developed the H-loss function and utilized it to advance the field of regression. The mathematical formulation of the H-loss can be presented as follows:
\begin{align}\label{HE Loss}
\mathcal{L}_{H}(x)=
\begin{cases}
\lambda \left[  1-\{-a(x+\varepsilon)+1\} e^{a(x+\varepsilon)}   \right], & x \leq -\varepsilon, \\
0, & -\varepsilon < x < \varepsilon,\\
\lambda \left[  1-\{a(x-\varepsilon)+1\} e^{-a(x-\varepsilon)}   \right], & x \geq \varepsilon,
\end{cases}
\end{align}
where \(\varepsilon > 0\) is an insensitivity parameter that creates a tolerance zone within the loss function, \(a > 0\) serves as a shape parameter, and \(\lambda > 0\) is a bounding parameter that constrains the loss to prevent it from exceeding a specified limit.
Figs. \ref{fig:first}, \ref{fig:second}, and \ref{fig:third} illustrate the H-loss function under various parameter settings. The H-loss function exhibits several desirable mathematical properties, including smoothness, boundedness, and symmetry, while simultaneously incorporating an insensitive zone. Now, we conduct a thorough analysis of these mathematical properties.
\begin{enumerate}
    \item \textbf{Smoothness:} The H-loss function is continuously differentiable across its entire domain. This implies that the first derivative, \(\frac{d\mathcal{L}_{\text{H}}}{dx}\), exists and is continuous, ensuring that the gradient of the loss function is well-behaved. Smoothness is crucial for gradient-based optimization algorithms, as it ensures stable and efficient convergence by avoiding abrupt changes in the loss landscape.
    \item \textbf{Symmetry:} The H-loss function is symmetric about the origin. Mathematically, this means that \(\mathcal{L}_{\text{H}}(-x) = \mathcal{L}_{\text{H}}(x)\). This ensures that the function treats positive and negative errors equally, maintaining balanced penalization across both sides of the error spectrum.
    \item \textbf{Boundedness:} The H-loss function is bounded, meaning that there exists a maximum value, \(\mathcal{L}_{\text{H}}^{\max}\), beyond which the loss cannot increase. This upper bound is particularly advantageous in the presence of outliers, as it prevents the loss from escalating uncontrollably due to extreme errors. Consequently, the model becomes more robust to outliers, maintaining stability and performance even in the presence of anomalous data points.
    \item \textbf{Insensitive zone:} The H-loss function features an insensitive zone around the origin, defined by the interval \(-\varepsilon < x < \varepsilon\), within which the loss is zero. This characteristic is beneficial for reducing the impact of small errors or noise in the data. By ignoring minor discrepancies, the insensitive zone prevents the model from overfitting to trivial variations, thereby improving the generalization capability of the model. This feature is particularly useful in practical applications where data often contains inherent noise or minor perturbations.
\end{enumerate}
The derivative of the H-loss function can be derived as follows:
\begin{align}\label{Derivate of HE Loss}
\frac{\partial\mathcal{L}_{H} (x)}{\partial x}=
\begin{cases}
\lambda a^2 (x+\varepsilon) e^{a(x+\varepsilon)}, & x < -\varepsilon, \\
0, & -\varepsilon \leq x \leq \varepsilon,\\
\lambda a^2 (x-\varepsilon) e^{-a(x-\varepsilon)}, & x > \varepsilon.
\end{cases}
\end{align}
Figs. \ref{fig:four} and \ref{fig:fifth} visually depict the derivative of \(\mathcal{L}_{\text{H}}(\cdot)\) for different values of $a$ and $\lambda$, respectively. The derivative's shape offers vital understanding into the behaviour of H-loss function to minimization via gradient descent or analogous optimization techniques. For any values of \(\lambda\) and \(a\) , the derivative is zero when \(|x| \leq \varepsilon\), indicating insensitivity to minor errors. However, as the error magnitude exceeds \(\varepsilon\), 
the derivative's steepness intensifies as a direct result of the increasing values of \(a\) and \(\lambda\). Moreover, as the error magnitude continues to rise beyond a certain threshold, the derivative's magnitude starts to decrease, ultimately approaching zero. This behavior implies that as the error magnitude of an outlier grows, its effect on the gradient descent process gradually weakens, and beyond a certain threshold, the outlier's influence becomes almost negligible.
\subsection{Formulation of H-RVFL}
In this subsection, we embed the H-loss function into RVFL and design a robust RVFL model for binary classification, designated as H-RVFL. We then employ the NAG algorithm to optimize the H-RVFL model, exploiting the smoothness of the H-loss function to achieve efficient and reliable training. The formulation of the proposed H-RVFL can be presented as:
\begin{align} \label{H-RVFL formulation}
   \underset{\hat{\beta}}{\text{argmin}}\,~ J(\hat{\beta}) =\frac{1}{2}\|\hat{\beta}\|_2^2 + \frac{C}{2} \sum_{i=1}^{n} \mathcal{L}_{H}(T_i \hat{\beta}-y_i),
\end{align}
where $C>0$ is a tunable parameter, $T$ denotes the concatenation matrix comprising the training data and output from the hidden layer, \(\mathcal{L}_{\text{H}}(\cdot)\) represents the H-loss function, $y$ denotes the label vector, and $\hat{\beta}$ denotes the output vector of dimension $(m+h_l \times 1)$.
The non-convex nature of the H-loss function makes the unconstrained optimization problem (\ref{H-RVFL formulation}) inherently non-convex, posing significant challenges in deriving a closed-form solution. Nevertheless, the smoothness characteristic of the H-loss function enables the use of gradient-based algorithms for effectively optimizing the H-RVFL model, despite the problem's non-convexity.
\par
The gradient of the optimization problem (\ref{H-RVFL formulation}) can be obtained as follows:
\begin{align} \label{Gradient H-RVFL}
\nabla J(\hat{\beta}_t)= \hat{\beta} + \frac{C}{2} \sum_{i=1}^{n}\frac{\partial\mathcal{L}_{H}}{\partial \hat{\beta}},
\end{align}
where
\begin{align*} \label{H-loss derivative}
\frac{\partial\mathcal{L}_{H}}{\partial \hat{\beta}}=
\begin{cases}
\lambda a^2 (\xi_i+\varepsilon)  e^{a(\xi_i+\varepsilon)} T_i^{\intercal}, & \xi_i < -\varepsilon, \\
0, & -\varepsilon \leq \xi_i \leq \varepsilon,\\
\lambda a^2 (\xi_i-\varepsilon) e^{-a(\xi_i-\varepsilon)} T_i^{\intercal}, & \xi_i > \varepsilon.
\end{cases}
\end{align*}
Here $T_i$ represents the $i^{th}$ row of the matrix $T$, $T_i^{\intercal}$ denotes the transpose of $T_i$, and $\xi_i = T_i \hat{\beta}-y_i$. 
\subsection{Optimization for H-RVFL}
To address the optimization problem (\ref{H-RVFL formulation}), we utilize the Nesterov accelerated gradient (NAG) algorithm. NAG is an advanced variant of stochastic gradient descent (SGD), designed to accelerate convergence by incorporating a momentum term. Momentum \cite{qian1999momentum} in gradient descent helps accelerate the optimization process by accumulating a velocity vector in directions of persistent gradient descent and dampening oscillations. NAG improves upon classical momentum by calculating the gradient at a look-ahead position.
There are two key reasons to prefer the NAG algorithm. 1) The look-ahead mechanism provides a more accurate update direction, leading to faster convergence compared to standard momentum methods; 2) By anticipating the future position, NAG reduces oscillations and provides smoother updates, which is particularly beneficial in ravine-like regions of the optimization landscape. 
Therefore, utilizing NAG to optimize the proposed H-RVFL model is both appropriate and beneficial.
\par
One challenge associated with NAG is selecting an optimal learning rate. If the learning rate is too low, the convergence speed of the algorithm becomes sluggish. On the other hand, a high learning rate can lead the algorithm to oscillate (overshoot the optimal point) or even fail to converge. A effective strategy is to initiate training with a moderately high learning rate and gradually decrease it according to a predetermined schedule. Inspired by the simulated annealing approach \cite{kirkpatrick1983optimization}, we adopt the exponential decay strategy to update the learning rate as \(\mu_{\text{new}} = \mu_{\text{old}} \exp(-\eta t)\), where \(\eta\) is a hyperparameter that govern the extent of the learning rate's decay at each iteration, and \(t\) represents the current iteration number. This strategy balances the trade-off between exploration and convergence speed, enhancing the overall efficiency of the NAG algorithm. Now, the step-by-step NAG algorithm to solve the H-RVFL model (\ref{H-RVFL formulation}) is as follows:
\begin{enumerate}
    \item \textbf{Initialization:} Initialize the model parameter \(\hat\beta_0\), velocity vector \(v_0\), learning rate \(\mu_0\). Set the momentum coefficient \(r\), learning rate decay parameter \(\eta\), and the maximum number of iterations \(I\).
\item \textbf{Iteration:} For each iteration \(t = 1, 2, \ldots, I\);
\begin{enumerate}
\item Compute look-ahead point:
\[ \hat{\beta}_{\text{look-ahead}} = \hat{\beta}_t + \gamma v_t. \]
\item Calculate the gradient at look-ahead point using equation (\ref{Gradient H-RVFL}):
     \[ g_t = \nabla J(\hat{\beta}_{\text{look-ahead}}).\] 
\item Update velocity vector:
      \[ v_{t+1} = r v_t - \mu_t g_t. \]
\item Update H-RVFL parameter:
     \[ \hat{\beta}_{t+1} = \hat{\beta}_t + v_{t+1}.\]
\item Adjust learning rate with exponential decay:
     \[ \mu_{t+1} = \mu_t \exp(-\eta t). \]
     \end{enumerate}
\item \textbf{Convergence check:} If the convergence criteria are met, terminate the iteration process. Common stopping criteria include reaching the maximum number of iterations or when the change in \(\hat{\beta}\) falls below a predefined threshold, indicating convergence.
\end{enumerate}
The structure of the NAG algorithm for solving the H-RVFL model is meticulously detailed in Algorithm \ref{NAG-algorithm-HRVFL}. Further, the computational complexity of the utilized NAG algorithm is provided in Section S.II of the supplement file.\\
Upon obtaining the optimal $\hat{\beta}$, the following decision function is employed to predict the label vector of the test set $(\tilde{X})$:
\begin{align*} \label{Decision function}
\tilde{y}= {\operatorname{\text{sign}}}\left( \tilde{T} \hat{\beta}\right),
\end{align*}
where $\tilde{T}$ is the concatenation matrix comprising the
test data and output from the hidden layer.

\begin{algorithm}
  \caption{NAG algorithm for Solving H-RVFL}
  \label{NAG-algorithm-HRVFL}
   \begin{algorithmic}
\STATE \textbf{Input:}
\STATE The dataset:  $\left\{x_i,y_i\right\}_{i=1}^n$.
\STATE The parameters: H-loss parameter $a$, $\lambda$, and $\varepsilon$; regularization parameter $C$; number of samples in mini-batch $k$; maximum iteration number $I$, learning rate decay factor $\eta$, momentum coefficient $r$, convergence tolerance $\theta$.
\STATE Initialize: $\hat{\beta}_0$, $v_0$ 
\STATE \textbf{Output:} $\hat{\beta}$;  
\STATE $1:$ Randomly select a mini-batch of \( k \) samples \(\{x_i, y_i\}_{i=1}^{k}\) uniformly.
\STATE $2:$  \textbf{while} $t \leq I$ 
\STATE $3:$  Look-ahead update: $\hat{\beta}_{\text{look-ahead}} = \hat{\beta_t} + \gamma v_t$;\\
\STATE $4:$ Compute the gradient $g_t$ using equation (\ref{Gradient H-RVFL});
\STATE $5:$  Update velocity vector: $v_{t+1} = r v_t - \mu_t g_t$;
\STATE $6:$  Update H-RVFL parameter: $\hat{\beta}_{t+1} = \hat{\beta}_t + v_{t+1}$;
\STATE $7:$  Update learning rate: $\mu_{t+1} = \mu_t \exp(-\eta t)$;
\STATE $8:$  ~~~~~\textbf{if }$\left\|\hat{\beta}_{t+1}-\hat{\beta}_t\right\|<\theta$ 
\STATE $9:$ ~~~~~~~\text{break}
\STATE $10:$  ~~~~~\textbf{else}
\STATE $11:$  ~~~~~~~$t \leftarrow t+1$
\STATE $12:$  ~~~~\textbf{end if}
\STATE $13:$  \textbf{end while}
\STATE $14:$  \textbf{Return} $\hat{\beta}$.
 \end{algorithmic}
\end{algorithm}


\section{Experimental Evaluation}
To validate the effectiveness of the proposed H-RVFL, we evaluate it on 40 UCI \cite{dua2017uci} and KEEL \cite{derrac2015keel} benchmark datasets (with and without label noise) across various domains. A detailed description of datasets is given in Table S.I of the supplement file. For comparison, we used 5 state-of-the-art models, namely SVM \cite{cortes1995support}, RVFL \cite{pao1994learning}, Extreme learning machine or RVFL without direct link (RVFLwoDL) \cite{huang2006extreme}, IF-RVFL \cite{malik2022alzheimer}, and NF-RVFL \cite{sajid2024neuro}. A detailed experimental setup with the hyperparameters setting is provided in Section S.III.A of the supplementary file.

\subsection{Experimental results and discussions on UCI and KEEL datasets}
In this subsection, we present the experimental outcomes obtained from 40 real-world datasets sourced from the UCI \cite{dua2017uci} and KEEL \cite{derrac2015keel} repositories. The datasets are categorized based on sample size into two groups: (Group A) datasets with $5000$ or fewer samples, and (Group B) datasets with more than $5000$ samples, up to a maximum of $130064$ samples. Group A comprises $33$ datasets, while Group B includes $7$ datasets.
Table \ref{tab:Average-UCI-KEEL-Table} presents the average performance comparison of the proposed H-RVFL model against five baseline models: SVM \cite{cortes1995support}, RVFLwoDL \cite{huang2006extreme}, RVFL \cite{pao1994learning}, NF-RVFL \cite{sajid2024neuro} and IF-RVFL \cite{malik2022alzheimer} on the 33 Group A datasets. The detailed results on each of the 33 (Group A) datasets are presented in Table S.II of the supplement file. The average accuracy of H-RVFL across all datasets in Group A is 80.76\%, which is higher than the accuracies of SVM (73.38\%), RVFLwoDL (75.17\%), RVFL (75.34\%), NF-RVFL (77.27\%), and IF-RVFL (78.61\%). 
This represents an approximate 2.15\% accuracy gain compared to the second-best model, IF-RVFL. The standard deviation of H-RVFL’s accuracy is also favorable, suggesting that the proposed model performance is stable and less variable across different datasets. The average standard deviation of H-RVFL’s accuracy is 7.31, which is notably lower than the standard deviations observed for several baseline models, such as SVM (15.77), RVFLwoDL (10.66), RVFL (10.31), and IF-RVFL (8.81). Although NF-RVFL has a slightly lower standard deviation of 6.63, H-RVFL achieves a higher average accuracy of 80.76\% compared to NF-RVFL’s 77.27\%, indicating superior stability and performance. The experimental results of the proposed H-RVFL model against the baseline model on 7 (Group B) large datasets are presented in Table \ref{tab:Large-UCI-KEEL-Table}. The proposed H-RVFL model achieves an impressive average accuracy of 76.05\%, significantly outperforming the baseline models. In contrast, the SVM, RVFLwoDL, RVFL, NF-RVFL, and IF-RVFL achieve an average accuracy of 58.84\%, 70.71\%, 71.13\%, 69.63\%, and 62.22\% respectively. It is important to note that the average accuracy of the SVM and IF-RVFL models is based on only three datasets (musk\_2, ringnorm, and twonorm), as both models encountered out-of-memory errors on four other datasets (adult, connect\_4, magic, and miniboone). This limitation underscores a significant practical challenge faced by traditional models when processing large and complex datasets. For SVM, the out-of-memory error is primarily due to the high memory consumption required to compute and store the kernel matrix. For IF-RVFL, the error is primarily attributed to the computational complexity involved in fuzzy membership calculation within a high-dimensional kernel space. In contrast, the proposed H-RVFL model not only outperforms the baseline models in terms of average accuracy but also demonstrates robustness and efficiency in handling large datasets without succumbing to memory constraints. 
\par
The average standard deviation of H-RVFL on 7 Group B datasets is 7.56. This is again lower compared to several baseline models, such as SVM (17.43), RVFLwoDL (13.65), RVFL (12.51), and IF-RVFL (8.51). While NF-RVFL shows a slightly lower standard deviation of 6.74, H-RVFL still outperforms it in terms of average accuracy (76.05\% versus 69.63\%). The low standard deviation of H-RVFL’s accuracy across both groups of datasets implies that the model performs consistently well, without significant fluctuations. Overall, the proposed H-RVFL model consistently outperformed the baseline models in terms of high accuracy and low variability. Its significant gains in accuracy demonstrate the effectiveness of incorporating the H-loss function and the NAG algorithm in improving the efficiency of the RVFL framework.
\par
To further affirm the effectiveness of the proposed H-RVFL model, a comprehensive statistical evaluation was conducted on 33 (Group A) UCI and KEEL datasets. This evaluation included three distinct analyses: a ranking test, the Nemenyi post hoc test, and the win-tie-loss sign test. A detailed analysis of the statistical results is discussed in Section S.III.B of the supplement file.
\begin{table}[t]
\setlength{\tabcolsep}{7pt}
\centering
\caption{Average performance comparison of the proposed H-RVFL with the baseline algorithms on 33 (Group A) datasets.}
\label{tab:Average-UCI-KEEL-Table}
\resizebox{\textwidth}{!}{%
\begin{tabular}{lcccccc}
\hline
 &  SVM \cite{cortes1995support}~~~  &  RVFLwoDL \cite{huang2006extreme}~~~  & RVFL \cite{pao1994learning}~~~ & NF-RVFL \cite{sajid2024neuro} & IF-RVFL \cite{malik2022alzheimer} &  H-RVFL$^{\dagger}$  \\
\textbf{Avg. Acc.$\pm$Avg.Std.}               & 73.38$\pm$15.77 & 75.17$\pm$10.66 & 75.34$\pm$10.31 & 77.27$\pm$6.63  & \underline{78.61$\pm$8.81}  & \textbf{80.76$\pm$7.31} \\
\hline
\textbf{Avg. Rank}      & 4.12  & 3.82  & 3.45  & 3.21  & \underline{2.39}  & \textbf{2.18}  \\
\hline
\multicolumn{7}{l}{$^{\dagger}$ denotes the proposed model.}\\
\multicolumn{7}{l}{Avg., Std., and Acc. represent average, standard deviation, and accuracy, respectively.}\\
\multicolumn{7}{l}{Boldface indicates the best models, while underline denotes the second-best models.}
\end{tabular}}
\end{table}

\begin{table}[t]
\setlength{\tabcolsep}{7pt}
\centering
\resizebox{\textwidth}{!}{
\caption{Performance comparison of the proposed H-RVFL with baseline algorithms on each of the 7 large (Group B) datasets.}
\label{tab:Large-UCI-KEEL-Table}
\begin{tabular}{lcccccc}
\hline
\hline
Dataset &  SVM \cite{cortes1995support}~~~  &  RVFLwoDL \cite{huang2006extreme}~~~  & RVFL \cite{pao1994learning}~~~ & NF-RVFL \cite{sajid2024neuro} & IF-RVFL \cite{malik2022alzheimer} &  H-RVFL$^{\dagger}$  \\
&
  Acc.$\pm$Std. &
  Acc.$\pm$Std. &
  Acc.$\pm$Std. &
  Acc.$\pm$Std. &
  Acc.$\pm$Std. &
  Acc.$\pm$Std. \\
\hline
adult      & * & 84.25$\pm$0.34  & 84.28$\pm$0.28  & 81.71$\pm$0.3   &*& 76.07$\pm$0.26  \\
connect\_4 & \multicolumn{1}{c}{*} & 75.39$\pm$5.9   & 75.38$\pm$5.9   & 75.09$\pm$5.51  &*& 75.38$\pm$5.9   \\
magic      & \multicolumn{1}{c}{*} & 76.65$\pm$19.73 & 76.62$\pm$19.91 & 77.38$\pm$10.16 &*& 89.86$\pm$20.28 \\
miniboone  & \multicolumn{1}{c}{*} & 72.17$\pm$36.78 & 75.22$\pm$28.71 & 83.13$\pm$3.81  &*& 96.93$\pm$6.12  \\
musk\_2    & 75$\pm$50                   & 84.59$\pm$30.82 & 84.59$\pm$30.82 & 67.37$\pm$24.77 &85.97$\pm$22.61& 90.41$\pm$19.18 \\
ringnorm   & 50.49$\pm$1.64              & 50.84$\pm$0.74  & 50.92$\pm$1.53  & 51.34$\pm$1.91  &51.03$\pm$1.42& 51.99$\pm$0.48  \\
twonorm    & 51.04$\pm$0.65              & 51.05$\pm$1.21  & 50.89$\pm$0.42  & 51.39$\pm$0.7   & 50.86$\pm$1.49& 51.7$\pm$0.68 \\
\hline
\textbf{Avg. Acc.$\pm$Avg. Std.}      & 58.84$\pm$17.43 & 70.71$\pm$13.65  & \underline{71.13$\pm$12.51}  & 69.63$\pm$6.74  & 62.62$\pm$8.51  & \textbf{76.05$\pm$7.56} 
\\
\hline
\multicolumn{7}{l}{$^{\dagger}$ denotes the proposed model.}\\
\multicolumn{7}{l}{$^{*}$ indicates that system encountered an out-of-memory error.}
\end{tabular}}
\end{table}
\begin{table}[t]
\setlength{\tabcolsep}{7pt}
\centering
\resizebox{\textwidth}{!}{
\caption{Performance comparison of the proposed H-RVFL with the baseline algorithms on 8 selected datasets with varying level of label noise.}
\label{tab:noise}
\begin{tabular}{lcccccccc}
 \hline 
 Models & &  SVM \cite{cortes1995support}~~~  &  RVFLwoDL \cite{huang2006extreme}~~~  & RVFL \cite{pao1994learning}~~~ & NF-RVFL \cite{sajid2024neuro} & IF-RVFL \cite{malik2022alzheimer} &  H-RVFL$^{\dagger}$  \\

 Dataset & Noise & Acc.$\pm$Std. &
  Acc.$\pm$Std. &
  Acc.$\pm$Std. &
  Acc.$\pm$Std. &
  Acc.$\pm$Std. &
  Acc.$\pm$Std. \\
\hline
 
breast\_cancer                   & 5\%                       & 70.14$\pm$47.65       & 70.14$\pm$47.65       & 70.14$\pm$47.65        & 66.52$\pm$25.87         & 70.14$\pm$47.65        & 95.49$\pm$9.03          \\
                                 & 10\%                      & 70.14$\pm$47.65       & 70.14$\pm$47.65       & 70.14$\pm$47.65        & 65.81$\pm$24.82         & 70.14$\pm$47.65        & 95.49$\pm$9.03          \\
                                 & 20\%                      & 70.14$\pm$47.65       & 70.14$\pm$47.65       & 70.14$\pm$47.65        & 65.45$\pm$25.46         & 70.14$\pm$47.65        & 95.49$\pm$9.03          \\
                                 & 30\%                      & 70.14$\pm$47.65       & 70.14$\pm$47.65       & 70.14$\pm$47.65        & 63.7$\pm$22.47          & 70.14$\pm$47.65        & 95.49$\pm$9.03          \\
                                 & 40\%                      & 70.14$\pm$47.65       & 70.14$\pm$47.65       & 70.14$\pm$47.65        & 67.53$\pm$25.75         & 70.14$\pm$47.65        & 95.49$\pm$9.03          \\
                                 \hline
\textbf{Avg. Acc.$\pm$Avg. Std.}                        & \multicolumn{1}{l}{}      & \underline{ 70.14$\pm$47.65} & \underline{ 70.14$\pm$47.65} & \underline{ 70.14$\pm$47.65}  & 65.8$\pm$24.87          & \underline{ 70.14$\pm$47.65}  & \textbf{95.49$\pm$9.03} \\
\hline
fertility                    & 5\%                       & 88$\pm$7.3            & 89$\pm$8.25           & 89$\pm$8.25            & 86$\pm$5.16             & 89$\pm$8.25            & 90$\pm$5.16             \\
                                 & 10\%                      & 88$\pm$7.3            & 88$\pm$7.3            & 89$\pm$8.25            & 84$\pm$12.65            & 89$\pm$8.25            & 90$\pm$5.16             \\
                                 & 20\%                      & 88$\pm$7.3            & 89$\pm$8.25           & 89$\pm$8.25            & 84$\pm$5.66             & 89$\pm$8.25            & 90$\pm$5.16             \\
                                 & 30\%                      & 88$\pm$7.3            & 89$\pm$8.25           & 89$\pm$8.25            & 87$\pm$8.87             & 89$\pm$8.25            & 90$\pm$5.16             \\
                                 & 40\%                      & 65$\pm$40.97          & 89$\pm$8.25           & 89$\pm$8.25            & 88$\pm$8.64             & 89$\pm$8.25            & 90$\pm$5.16             \\
                                 \hline
\textbf{Avg. Acc.$\pm$Avg. Std.}                       & \multicolumn{1}{l}{}      & 83.4$\pm$14.04        & 88.8$\pm$8.06         & \underline{ 89$\pm$8.25}      & 85.8$\pm$8.2            & \underline{ 89$\pm$8.25}      & \textbf{90$\pm$5.16}    \\
\hline
haberman\_survival           & 5\%                       & 73.51$\pm$4.03        & 73.51$\pm$4.03        & 73.51$\pm$4.03         & 76.46$\pm$1.99          & 73.51$\pm$4.03         & 74.5$\pm$3.89           \\
                                 & 10\%                      & 73.51$\pm$4.03        & 73.51$\pm$4.03        & 73.51$\pm$4.03         & 73.51$\pm$4.03          & 73.51$\pm$4.03         & 74.82$\pm$3.89          \\
                                 & 20\%                      & 73.51$\pm$4.03        & 73.51$\pm$4.03        & 73.51$\pm$4.03         & 74.5$\pm$3.2            & 73.51$\pm$4.03         & 75.15$\pm$4             \\
                                 & 30\%                      & 74.18$\pm$4.37        & 73.84$\pm$4.65        & 73.84$\pm$4.65         & 74.49$\pm$4.51          & 73.84$\pm$4.65         & 74.82$\pm$3.89          \\
                                 & 40\%                      & 73.51$\pm$4.03        & 73.51$\pm$4.03        & 73.51$\pm$4.03         & 74.82$\pm$4.07          & 73.51$\pm$4.03         & 74.49$\pm$3.93          \\
                                 \hline
\textbf{Avg. Acc.$\pm$Avg. Std.}                       & \multicolumn{1}{l}{}      & 73.65$\pm$4.1         & 73.58$\pm$4.16        & 73.58$\pm$4.16         & \textbf{74.76$\pm$3.56} & 73.58$\pm$4.16         & \underline{ 74.75$\pm$3.92}    \\
\hline
hepatitis                    & 5\%                       & 79.25$\pm$14.08       & 83.89$\pm$8.17        & 84.53$\pm$8.6          & 83.16$\pm$7.08          & 84.53$\pm$8.6          & 82.46$\pm$13.26         \\
                                 & 10\%                      & 79.25$\pm$14.08       & 82.54$\pm$8.36        & 83.22$\pm$3.38         & 82.47$\pm$12.66         & 83.22$\pm$3.38         & 83.1$\pm$13.46          \\
                                 & 20\%                      & 79.25$\pm$14.08       & 83.84$\pm$10.21       & 83.87$\pm$6.73         & 80.6$\pm$7.88           & 83.87$\pm$6.73         & 82.46$\pm$13.26         \\
                                 & 30\%                      & 79.25$\pm$14.08       & 83.18$\pm$6.66        & 81.88$\pm$7.78         & 76.1$\pm$13.06          & 81.88$\pm$7.78         & 83.1$\pm$13.46          \\
                                 & 40\%                      & 79.25$\pm$14.08       & 81.87$\pm$9.41        & 81.85$\pm$12.49        & 74.83$\pm$6.44          & 81.85$\pm$12.49        & 82.46$\pm$13.26         \\
                                 \hline
\textbf{Avg. Acc.$\pm$Avg. Std.}                       & \multicolumn{1}{l}{}      & 79.25$\pm$14.08       & \underline{ 83.06$\pm$8.56}  & \textbf{83.07$\pm$7.8} & 79.43$\pm$9.43          & \textbf{83.07$\pm$7.8} & 82.71$\pm$13.34         \\
\hline
ilpd\_indian\_liver          & 5\%                       & 71.36$\pm$5.58        & 71.36$\pm$5.58        & 71.36$\pm$5.58         & 69.47$\pm$5.47          & 71.36$\pm$5.58         & 71.53$\pm$5.8           \\
                                 & 10\%                      & 71.36$\pm$5.58        & 71.36$\pm$5.58        & 71.36$\pm$5.58         & 68.78$\pm$5.63          & 71.36$\pm$5.58         & 71.53$\pm$5.69          \\
                                 & 20\%                      & 71.36$\pm$5.58        & 71.36$\pm$5.58        & 71.53$\pm$5.53         & 70.15$\pm$4.52          & 71.53$\pm$5.53         & 71.53$\pm$5.69          \\
                                 & 30\%                      & 71.36$\pm$5.58        & 71.36$\pm$5.58        & 71.36$\pm$5.58         & 70.5$\pm$5.03           & 71.36$\pm$5.58         & 71.87$\pm$4.65          \\
                                 & 40\%                      & 71.36$\pm$5.48        & 71.36$\pm$5.58        & 71.53$\pm$6.16         & 71.7$\pm$3.12           & 71.53$\pm$6.16         & 71.87$\pm$4.65          \\
                                 \hline
\textbf{Avg. Acc.$\pm$Avg. Std.}                       & \multicolumn{1}{l}{}      & 71.36$\pm$5.56        & 71.36$\pm$5.58        & \underline{ 71.43$\pm$5.68}   & 70.12$\pm$4.76          & \underline{ 71.43$\pm$5.68}   & \textbf{71.67$\pm$5.3}  \\ 
\hline
pittsburg\_bridges\_T\_OR\_D & 5\%                       & 86.15$\pm$14.87       & 86.15$\pm$14.87       & 86.15$\pm$14.87        & 87.19$\pm$7.66          & 86.15$\pm$14.87        & 90.15$\pm$6.99          \\
                                 & 10\%                      & 86.15$\pm$14.87       & 87.12$\pm$16.14       & 86.15$\pm$14.87        & 81.31$\pm$15.19         & 86.15$\pm$14.87        & 90.15$\pm$6.99          \\
                                 & 20\%                      & 86.15$\pm$14.87       & 86.15$\pm$14.87       & 86.15$\pm$14.87        & 82.23$\pm$12.16         & 86.15$\pm$14.87        & 90.15$\pm$6.99          \\
                                 & 30\%                      & 86.15$\pm$14.87       & 86.15$\pm$14.87       & 86.15$\pm$14.87        & 83.27$\pm$13.69         & 86.15$\pm$14.87        & 88.19$\pm$8.69          \\
                                 & 40\%                      & 86.15$\pm$14.87       & 86.15$\pm$14.87       & 86.15$\pm$14.87        & 83.27$\pm$13.69         & 86.15$\pm$14.87        & 89.15$\pm$8.94          \\
                                 \hline
\textbf{Avg. Acc.$\pm$Avg. Std.}                       & \multicolumn{1}{l}{}      & 86.15$\pm$14.87       & \underline{ 86.35$\pm$15.12} & 86.15$\pm$14.87        & 83.45$\pm$12.48         & 86.15$\pm$14.87        & \textbf{89.56$\pm$7.72} \\
\hline
planning                     & 5\%                       & 72.48$\pm$7.13        & 71.38$\pm$7.69        & 71.38$\pm$7.69         & 68.7$\pm$5.95           & 71.38$\pm$7.69         & 73.57$\pm$6.75          \\
                                 & 10\%                      & 71.38$\pm$7.69        & 71.38$\pm$7.69        & 71.38$\pm$7.69         & 68.1$\pm$4.11           & 71.38$\pm$7.69         & 73.56$\pm$8.5           \\
                                 & 20\%                      & 71.38$\pm$7.69        & 71.38$\pm$7.69        & 71.38$\pm$7.69         & 61.53$\pm$3.84          & 71.38$\pm$7.69         & 74.12$\pm$7.53          \\
                                 & 30\%                      & 71.38$\pm$7.69        & 71.38$\pm$7.69        & 71.38$\pm$7.69         & 71.39$\pm$5.06          & 71.38$\pm$7.69         & 74.12$\pm$7.53          \\
                                 & 40\%                      & 71.39$\pm$6.22        & 71.38$\pm$7.69        & 71.38$\pm$7.69         & 69.73$\pm$6.86          & 71.38$\pm$7.69         & 73.03$\pm$6.07          \\
                                 \hline
\textbf{Avg. Acc.$\pm$Avg. Std.}                        & \multicolumn{1}{l}{}      & \underline{ 71.6$\pm$7.29}   & 71.38$\pm$7.69        & 71.38$\pm$7.69         & 67.89$\pm$5.16          & 71.38$\pm$7.69         & \textbf{73.68$\pm$7.28}\\
\hline
spambase & 5\%                  & 77.49$\pm$11.14 & 84.64$\pm$7.11  & 86.48$\pm$6.51  & 87.39$\pm$4.16  & 86.48$\pm$6.51  & 92.91$\pm$13.77          \\
             & 10\%                 & 77.36$\pm$12.29 & 85.77$\pm$7.11  & 86.77$\pm$7.18  & 87.2$\pm$5.18   & 86.77$\pm$7.18  & 91.83$\pm$16.35          \\
             & 20\%                 & 76.92$\pm$13.81 & 85.4$\pm$8.94   & 87.29$\pm$9.19  & 87.94$\pm$4.33  & 87.29$\pm$9.19  & 92.39$\pm$13.72          \\
             & 30\%                 & 75.94$\pm$15.37 & 84.09$\pm$12.67 & 85.79$\pm$12.09 & 87.29$\pm$3.1   & 85.79$\pm$12.09 & 92.52$\pm$14.96          \\
             & 40\%                 & 75.36$\pm$9.28  & 84.31$\pm$13.54 & 85.51$\pm$12.9  & 85.59$\pm$12.74 & 85.51$\pm$12.9  & 91.09$\pm$17.65          \\
\hline
\textbf{Avg. Acc.$\pm$Avg. Std.}    & \multicolumn{1}{l}{} & 76.61$\pm$12.38 & 84.84$\pm$9.88  & 86.37$\pm$9.57  & \underline{87.08$\pm$5.9}   & 86.37$\pm$9.57  & \textbf{92.15$\pm$15.29} \\
\hline
\multicolumn{7}{l}{$^{\dagger}$ denotes the proposed model.}\\
\multicolumn{7}{l}{Avg., Std., and Acc. represent average, standard deviation, and accuracy, respectively.}\\
\multicolumn{7}{l}{Boldface indicates the best models, while underline denotes the second-best models.}
\end{tabular}}
\end{table}
\subsection{Experimental results and discussions on UCI and KEEL datasets with label noise}
Real-world datasets like UCI and KEEL provide realistic circumstances for evaluating the performance of the proposed H-RVFL model; however, it is essential to recognize that label noise can occur in various scenarios, such as manual data entry errors, ambiguous or subjective labeling, and inherent inconsistencies in data collection processes.
To demonstrate the robustness of the proposed H-RVFL models under challenging conditions, we deliberately added label noise into a selection of datasets. Eight diverse datasets were chosen for this evaluation, including breast\_cancer, fertility, haberman, hepatitis, ilpd\_indian\_liver, pittsburg\_bridges\_T\_OR\_D, planning, and spambase. The labels of these datasets were corrupted with noise at different levels of 5\%, 10\%, 20\%, 30\%, and 40\% to simulate different degrees of label inaccuracies and assess the models' performance.
The performance results of the proposed H-RVFL model, compared to baseline models with varying noise levels, are detailed in Table \ref{tab:noise}. The performance of each model is evaluated using accuracy and standard deviation. The average accuracy and standard deviation across all noise levels for each dataset is also calculated to provide a comprehensive performance summary. Among the eight datasets, the H-RVFL model demonstrated superior average accuracy in six instances. Additionally, it achieved second and third place in one instance each, underscoring its robustness and efficacy in handling label noise. Moreover, the proposed H-RVFL model exhibited the lowest average standard deviation in five out of the eight datasets, indicating greater stability compared to other baseline algorithms. Hence, the proposed H-RVFL model demonstrated significant robustness and stability across varying levels of label noise. This robustness can be attributed to the incorporation of the H-loss function, which effectively mitigates the impact of noise and outliers, enhancing the model's overall performance.
\section{Conclusions}
In conclusion, we presents the H-RVFL model, an enhanced version of the RVFL framework designed to address the sensitivity to outliers and noise typically observed in traditional RVFL models. By embedding the HawkEye loss (H-loss) function, which offers boundedness, smoothness, and an insensitive zone, the H-RVFL model significantly improves robustness and efficiency in handling noisy datasets. Furthermore, the application of the Nesterov accelerated gradient (NAG) algorithm effectively addressed the non-convex optimization problem associated with the proposed H-RVFL model, ensuring efficient and stable convergence during training. The experimental evaluations on a diverse set of benchmark datasets (with and without label noise) demonstrate the superior performance of the proposed H-RVFL model compared to baseline models.
\par
Notably, this paper initiates the use of a bounded non-convex loss function in the RVFL framework. In future work, one can explore the incorporation of the H-loss function into advanced deep RVFL structures, potentially unlocking further improvements in model robustness and performance across various applications.

\section*{Acknowledgment}
This project receives support from the Science and Engineering Research Board through the Mathematical Research Impact-Centric Support (MATRICS) scheme, with Grant No. MTR/2021/000787. Mohd. Arshad receives funding support from SERB under the Core Research Grant (CRG/2023/001230). Additionally, Mushir Akhtar's research fellowship is funded by the Council of Scientific and Industrial Research (CSIR), New Delhi, under Grant No. 09/1022(13849)/2022-EMR-I.

\bibliography{refs.bib}
\bibliographystyle{abbrvnat}
\end{document}